\documentclass[conference]{IEEEtran}

\usepackage{times}
\usepackage{multirow}
\usepackage{cite}
\usepackage{amsmath,amssymb,amsfonts,url}
\usepackage{algorithm}
\usepackage{algorithmic}
\usepackage{graphicx}
\usepackage{textcomp}
\usepackage{xcolor}
\PassOptionsToPackage{dvipsnames}{xcolor}
\usepackage[font=footnotesize]{caption}

\usepackage{caption}
\usepackage{booktabs}

\usepackage[colorlinks=true,linkcolor=red,breaklinks=True,citecolor=brown]{hyperref}
\PassOptionsToPackage{hyphens}{url}

\def\BibTeX{{\rm B\kern-.05em{\sc i\kern-.025em b}\kern-.08em
    X_T\kern-.1667em\lower.7ex\hbox{E}\kern-.125emX}}
\begin{document}

\title{Concept-based Adversarial Attacks: Tricking Humans and Classifiers Alike}

\author{\IEEEauthorblockN{Johannes Schneider, Giovanni Apruzzese}
 \IEEEauthorblockA{\textit{Institute of Information Systems -- University of Liechtenstein} \\
 \{johannes.schneider, giovanni.apruzzese\}@uni.li
 }}

\maketitle

\begin{abstract}
We propose to generate adversarial samples by modifying activations of upper layers encoding semantically meaningful concepts. The original sample is shifted towards a target sample, yielding an adversarial sample, by using the modified activations to reconstruct the original sample. A human might (and possibly should) notice differences between the original and the adversarial sample. Depending on the attacker-provided constraints, an adversarial sample can exhibit subtle differences or appear like a ``forged'' sample from another class.
Our approach and goal are in stark contrast to common attacks involving perturbations of single pixels that are not recognizable by humans. Our approach is relevant in, e.g., multi-stage processing of inputs, where both humans and machines are involved in decision-making because invisible perturbations will not fool a human. 
Our evaluation focuses on deep neural networks. We also show the transferability of our adversarial examples among networks.
\end{abstract}


\section{Introduction}
\label{sec:introduction}
Adversarial samples are typically generated to be non-recognizable by humans\cite{akh21,su2019one}. This is commonly achieved by adding perturbations resulting from a specific optimization process to the inputs.
While this is arguably a preferred adversarial scenario, in this work, we aim at a different objective: generating adversarial samples by modifying ``higher-level concepts''. This leads to adversarial samples that are, possibly with some effort, recognizable through human inspection. 

An attacker might aim to construct adversarial samples that are not only classified differently from their ground truth but also contain specific attributes and are therefore perceived differently by a human. This likely holds in a multi-stage classification process with a human in the loop. For example, a seller might list an item for sale on an online marketplace (e.g. a house or a car), and upload images or other information on such item. Such a marketplace might employ existing artificial intelligence (AI) systems relying on deep learning (DL) to obtain a price estimate\cite{var18}. An attacker, i.e., a malicious seller, may want to simultaneously trick (i) such an AI system to output higher price estimates, and (ii) potential buyers by making them believe that the price is appropriate. The attacker might want to enhance visual aspects, increasing the item's appeal (e.g., making a house appear under lighting conditions impossible in reality or employing perspective distortions). Thus, the optimization objective for an attacker might be to change a given (original) sample towards a target sample (e.g., an image of an expensive house with appealing lighting conditions). Put differently, adversarial and original samples tend to exhibit visible differences that are likely to aim to deceive or nudge a human observer---a theme that recently attracted much attention (e.g., \cite{sch20dec}). The predicted class of the resulting adversarial sample should generally be identical to the class of the target sample but could be any class. The adversarial sample should also contain specific characteristics (also called ``concepts'') of the target sample---at least to some extent, since we constrain the adversarial sample to be similar to the given (original) sample. For example, consider an online seller of fashion articles whose website is crawled by a search engine classifying images into brands. The fashion seller might want to fool humans by altering the images of its articles. Potential buyers should perceive it as more appealing but should not complain that the shown product differs from the actual physical product. Aside from tricking humans, the seller might also aim to mislead an AI of the search engine, i.e., conduct search engine poisoning\cite{leo14}. The search engine might utilize image recognition to identify brands of items. The seller might want the search engine to confuse the image of his/her product with that of a well-known, commonly searched, expensive brand. 

 
Optimizing for non-recognizability by humans might not necessarily be the best option to disguise an adversarial sample if humans check (some) inputs and outputs of an AI to detect attacks. Misclassifications due to imperceptible adversarial perturbations (e.g.,~\cite{su2019one}) are ``astonishing'' for humans, who will likely suspect that an adversarial attack is taking place and will quickly react (e.g., sanitizing the input, applying defenses, or re-training the AI system). However, if the misclassified sample looked somewhat confusing, though still well-recognizable, a human is less likely to suspect an attack. Legal consequences can differ if such a sample could occur in reality. In this case, it is difficult to prove that the sample was created to deceive the classifier, since altering depictions of articles are common practice (in marketing). Rather the AI system might be said to perform poorly. 

This paper discusses how to generate adversarial samples by modifying higher-level representations of inputs. We focus on how to create adversarial samples similar to a given target or desired sample, i.e., exhibiting characteristics that should be present in the adversarial sample. 
Technically, our work is based on encoding and decoding between different spaces, i.e., a space derived from representations of classification models, a space derived from visual representation as perceived by humans, and linear interpolation. To demonstrate the effectiveness of our method, we evaluate common DL models against our adversarial examples and assess their transferability across different DL models.


\section{Background and Related Work} \label{sec:related}
Machine Learning (ML) represents a valuable asset for modern tasks~\cite{jordan2015machine}; in particular, Deep Learning has become a true technology enabler~\cite{lecun2015deep}. However, many efforts have shown that all such methods are vulnerable to \textit{adversarial attacks} (e.g.,~\cite{biggio2013evasion}), giving birth to a new research area commonly referred to `adversarial machine learning'~\cite{biggio2018wild}. Let us summarize this field and outline the unique traits of our paper.

\subsection{Adversarial Machine Learning}
Attacks against ML leverage the so-called `adversarial samples': the principle is to apply a perturbation to an input sample so as to trick an ML model into producing an incorrect output~\cite{papernot2018sok}. Previous efforts showed that similar attacks could break any ML model, including those based on traditional classifiers (e.g., SVM~\cite{pierazzi2020intriguing}) and DL ones (e.g.,~\cite{su2019one}). Noteworthy is also that adversarial attacks can be successful in diverse domains (and, hence, on different data types), such as computer vision~\cite{su2019one}, speech analysis~\cite{cisse2017houdini}, or cyber detection~\cite{apruzzese2019addressing}. Even \textit{real} ML systems have been defeated (e.g.,~\cite{liang2016cracking, wu2020making}). 

Adversarial attacks are denoted with a \textit{threat model} describing the relationship of the attacker with the target system, usually in the form of \textit{goal}, \textit{knowledge}, \textit{capability} and \textit{strategy}~\cite{biggio2018wild}. Common terms associated to such threat models are `white-box' and `black-box'~\cite{papernot2018sok}: the former envision attackers with complete knowledge of the target ML system, whereas in the latter the attacker knows nothing. Depending on such assumptions, the attacker can opt for different strategies, such as crafting the `perfect' perturbation---e.g., the well-known `Projected Gradient Descent' (PGD) method~\cite{madry2018towards}---or exploit the \textit{transferability} property of adversarial examples~\cite{demontis2019adversarial}. Although all adversarial attacks share the same objective (i.e., compromising the decisions of a ML system), the attacker may have more specific goals. Such goals can include minimizing the perturbations~\cite{su2019one}, reducing the amount of interactions with the target system~\cite{papernot2017practical}, inducing a `targeted' output~\cite{carlini2018audio}; but also more advanced goals are possible (e.g.~\cite{zhang2021seat}).

With respect to previous works, we consider a unique threat model: such uniqueness is given by the different \textit{goals} and \textit{strategies} of our adversarial samples. Indeed, we assume that the attacker wants to introduce perturbations that \textit{can be spotted by a human}; to do so, the attacker leverages \textit{high-level concepts} provided by the first layers of deep neural networks. This differs considerably from previous work (e.g.,~\cite{su2019one, moosavi2016deepfool}).  
Our attacks can be considered as introducing perturbations during the ``preprocessing'' phase of an AI system (e.g.,~\cite{quiring2020adversarial, xiao2019seeing}); however, our proposal differs in the \textit{method} used to generate such perturbations. Indeed, we leverage techniques (i.e.,~\cite{sch21cla}) within the \textit{explainable AI} (XAI) domain~\cite{bhatt2020explainable}: some papers use similar techniques to `explain' adversarial attacks (e.g.,~\cite{amich2021explanation}), but no paper uses such techniques to `generate' adversarial samples---to the best of our knowledge.

In particular, our perturbations involve techniques (i.e.,~\cite{sch21cla,sch22b}) within the \textit{explainable AI} (XAI) domain~\cite{bhatt2020explainable}: some papers use similar techniques to `explain' adversarial attacks (e.g.,~\cite{amich2021explanation}), but no paper uses such techniques to `generate' adversarial samples---to the best of our knowledge. 
Our work also relates to natural adversarial samples or out-of-distribution samples~\cite{hen21,rob20}, since our resulting samples can look natural but different from typical, frequent samples.

\subsection{Artificial Sample Generation}
Synthtetically generating samples from latent spaces, as done in this work, is a rich topic in research. It has been extremely successful in the last few years using generative adversarial networks~\cite{wan21} as well as variational autoencoders which can (re)construct images from a latent space. In creative domains, e.g., the arts, the latent representation used to generate new samples is often randomly chosen~\cite{schne22c}. In contrast, we focus on latent representations originating from actual samples or interpolations between their latent representations. In security, outputs of the last layer of ML models have been used to reconstruct training samples~\cite{fred15}. Their the goal is not to generate novel samples but rather to retrieve as much information on a specific training sample as possible. Traversing the latent space to generate samples between two classes is commonly used to understand the decisions of classifiers, which is relevant to XAI. For instance, the authors of~\cite{van20} construct contrastive explanations by using a variational autoencoder, and a linear SVM---the latter served as a separate classifier from the model to understand. They traversed the latent space feeding latent codes into a CNN to assess when a sample changed its class and can be considered contrastive. With respect to~\cite{van20} we assume a different setting. Specifically, we do not use a linear SVM (but rather parts of the model to be attacked). We potentially alter (given) original and target samples to be assessed before embedding them into the latent space, i.e., we consider multiple embedding spaces---the auto-encoding space, and the space of the classifier, i.e., activations of layer $L_t$.

%

\section{Problem Definition} \label{sec:problem}

\subsection{Requirements}
Our method requires complete knowledge of a DL model $M=(L_0,L_1,...,L_{k-1})$ being a sequence of $k$ layers, a target layer $L_t \in M$, an original sample $X_O$ and a dataset $\mathcal{X}=\{X\}$, i.e., (input) samples that stem from the data distribution the model should handle, e.g., training data without any label information. We denote by $M_{\text{-}t}$ and $M_{t\text{-}}$ the submodel consisting of the first layers up to layer $t$ (including it) and from $L_{t+1}$ up to the last layer $L_{k-1}$ (including it).  The output of the model $M(X)$ is a probability $p(Y|X)$ for each class $Y$, the output of a model $M_{\text{-}t}(X)$ are the activations of layer $L_t$ for the input $X$. We denote $Y^G_X$ as the ground truth class of $X$ and $Y^M_X$ as the predicted class by model $M$. Such settings can represent either `white-box' (if $M$ is the attacked model) or a `black-box' (if $M$ is a surrogate of the attacked model~\cite{papernot2017practical}) assumptions: as we will show (§\ref{sec:evaluation}), our adversarial examples can transfer between similar $M$. Finally, we assume an attacker that cannot\footnote{This is a \textit{realistic} assumption~\cite{apruzzese2021modeling}: an attacker with write-access to $\mathcal{X}$ would \textit{poison} such dataset (e.g.,~\cite{shafahi2018poison}) instead of using our proposed method.} affect the training procedure of $M$. 

\subsection{Objectives}
We consider three objectives to construct an adversarial sample $X_A$ based on the original sample $X_O$ for a given target sample $X_T$.
\begin{enumerate}
    \item \emph{Being ``mis-classified''}: The adversarial sample should be classified as the class of $X_T$, i.e., $Y^M_{X_A}=Y^G_{X_T}$.
    
    \item \emph{Mimicking (concepts of) a target sample}: The adversarial sample $X_A$ should be similar to $X_T$ as measured by a given encoding function $f$, i.e., the L2-norm $||f(X_A)-f(X_T)||$ should be small. The encoding function $f$ might measure semantic differences by transforming to the feature space of the classifier using $f(X):=M_{\text{-}t}(X)$, the identity function $f(X):=X$ or an encoder (of an autoencoder trained on $\mathcal{X}$) or, possibly, $f$ is given by human judgment assessing to what extent $X_O$ and $X_A$ are similar with respect to pre-defined characteristics. The identity function $f(X)=X$ constrains visual differences in image space (recognizable by humans), while $f(X):=M_{\text{-}t}(X)$ constrains differences of information relevant to the classifier. It can happen that two samples $X, X'$ appear identical for the classifier (at a higher abstraction level), i.e. $||M_{\text{-}t}(X)-M_{\text{-}t}(X')||=0$, although they appear very differently for a human. We use the identity function $f(X)=X$, i.e., $||X_O-X_A||<\eta$. That is, we constrain differences in image space. 
    
    \item \emph{Remaining similar to the original sample}: The adversarial sample $X_A$ should be close to the original, i.e., $||f(X_O)-f(X_A)||<\eta$ for a given $\eta$ and encoding function $f$. We consider the same encoding function $f$ as described priorly.

\end{enumerate}
%
Additionally, one might require that $X_A$ is a sample of the distribution the model $M$ was trained for. Assume $P$ provides the likelihood of a data sample then we might aim for $\max P(X_A)$. The likelihood function $P$ could be approximated using the training data. Our method perturbing only upper layer activations tends to produce ``likely'' samples $X_A$, i.e., samples often (but not always) appear to be realistic and they do not contain ``invisible'' (unnatural) modifications that trigger a misclassification (unlike traditional attacks). 
 


\begin{figure*}
\vspace{-6pt}
  \centering
  \includegraphics[width=0.9\linewidth]{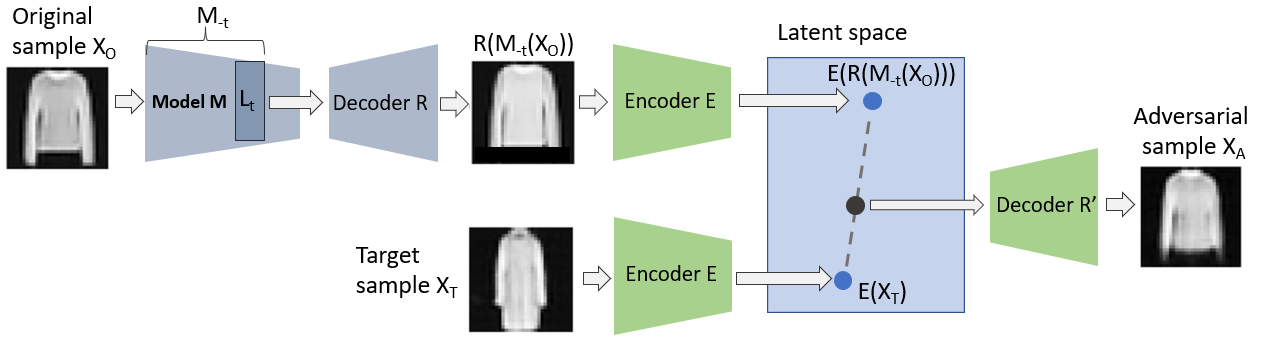}
  \caption{Semantic attack using an original sample $X_O$ and a target sample $X_T$. We first remove information not relevant for classification from the original sample through encoding and decoding before reconstructing an interpolation between the autoencoded representation of the original and the target sample.} \label{fig:over}
\end{figure*}

\section{Method}
\label{sec:method}
Our method aims to compute an adversarial sample $X_A$ that bears semantic similarity to a target sample $X_T$, while obeying constraints on the maximum perturbation. One way to achieve this is to linearly interpolate between the original sample $X_O$ and the target sample $X_T$, i.e. $X_A:=b\cdot X_O + (1-b)\cdot X_T$ by minimizing $b \in [0,1]$ given the constraint $||X_O-X_A||<\eta$. To find a (locally) optimal coefficient $b$ binary search can be used as stated in Algorithm \ref{alg:li}. 

\begin{algorithm}[htp]
	\caption{Linear interpolation with constraints\\
	\textbf{Input:} Target sample $X_T$, original sample $X_O$, encoder $E$, decoder $R'$, max. difference $\eta$   \\
	\textbf{Output:} Adversarial sample $X_A$}  \label{alg:li}
	\begin{algorithmic}[1]
		    \STATE $b_0:=0;b_1=1$ \COMMENT{value range for linear interpolation}
		    \STATE $\epsilon:=1e-2$ \COMMENT{Stopping criterion, i.e., precision of linear interpolation coefficient}
		    \WHILE{$b_1-b_0>\epsilon$}
		        \STATE $b:=\frac{b_0+b_1}{2}$
		        \STATE $E(X_A):=b\cdot E(X_O)+(1-b)\cdot E(X_T)$ \COMMENT{encoding of adversarial sample}
		        \STATE $X_A:=R'(E(X_A))$ \COMMENT{(decoded) adversarial sample}
		        \STATE \textbf{if } $||X_O-X_A||<\eta$ \textbf{ then } $b_0:=b$
		        \STATE \textbf{else } $b_1:=b$
		    \ENDWHILE
		    \STATE return $X_A$
	\end{algorithmic}	
\vspace{-2pt}
\end{algorithm}

As we shall discuss (§\ref{sec:qua}), the outcome of linear interpolation of samples tends to be easily recognizable as being artificially constructed. Therefore, one might perform the interpolation in a latent space given by an encoder $E$ stemming from an autoencoder $AE=(E,R')$, i.e., $E(X_A):=b\cdot E(X_O) + (1-b)\cdot E(X_T)$.  The $AE$, i.e., the encoder $E$ and decoder $R'$ are trained on training data $\mathcal{X}$. The adversarial sample can be obtained using the decoder, i.e., $X_A:=R'(E(X_A))$. However, we propose to interpolate from an input $X_O'$ of $X_O$ that contains only information on $X_O$ captured at layer $L_t$, i.e., by $M_{-t}(X_O)$. We obtain $X_O'$ using a decoder $R$, i.e., $X_O'=R(M_{-t}(X_O))$.  The decoder $R$ is trained by using activations $M_{-t}(X)$ as inputs and reconstructing $X$, i.e., the decoder $R$ minimizes $||X-R(M_{-t}(X))||$ for $X \in \mathcal{X}$~\cite{sch21cla,sch22b}. 
The process to obtain an adversarial sample $X_A$ is illustrated in Fig.~\ref{fig:over}. To understand the process, conceptually, we can think of two spaces: a classifier embedding space, and a latent embedding space from the autoencoder. To alter the class of a sample we can move in either space and use a decoder to get a sample based on the latent representations. When using the classifier embedding space, specific concepts of the original sample (and target sample) might be ignored, because they are not relevant for classification as discussed in the context of explainability~\cite{sch21cla,sch22b}. In our work, we also aim to obtain samples similar to a given target sample. Thus, linearly interpolating between encodings of the original sample and the target sample in classifier space seems not the right way, since the encoding of the target sample is stripped of information that we do not necessarily want to discard. Our objective is to be visually similar to the target sample. It seems more appealing to interpolate in an autoencoding space which allows to accurately reconstruct samples. A viable approach is to either use the original sample's encoding directly, or an encoding of the original sample.
Samples with irrelevant information or rare characteristics of a sample tend to be poorly reconstructed. Their reconstruction can appear ``blurry'' resembling more of an average. For example, if a T-shirt has a very unique pattern, shape or color, none of which is very relevant for classification, the reconstruction might fail to reconstruct it accurately, resulting in a blurry outline (shape), ``average'' pattern (e.g., uniform color), and an average color (e.g., grey tone) \cite{sch22b}. Thus, generally, the reconstruction appears more of an average. Therefore, common samples close to the average will be reconstructed better, whereas `outliers' are poorly decoded.
Overall, removing non-classification relevant information from $X_O$, i.e., using $X_O'=R(M_{-t}(X_O))$, pushes $X_O$ to denser areas, where reconstructions are better, i.e., difference $||X_A - X_O||$ is lower, and it is not negatively impacting the objective to maximize, i.e., $p(Y^G_D|X_A)$.

\section{Evaluation}
\label{sec:evaluation}
We perform a qualitative and quantitative evaluation focusing on image classification using CNNs. We assess multiple methods based on linear interpolation with the target sample $X_T$ and original sample $X_O$ embedded in a (latent) space with appropriate reconstruction. We investigate four scenarios using linear interpolation:
\begin{itemize}
    \item No encoding: Interpolation from $X_O$ to $X_T$ in image space.
    \item Classifier encoding $M_{\text{-}t}$: Interpolation from $M_{\text{-}t}(X_O)$ to $M_{\text{-}t}(X_T)$ in space given by $M_{\text{-}t}$; as decoder we use $R$;
    \item Encoder $E$ from the autoencoder $AE\!=\!(E,R')$ and a version of $X_O$ maintaining information relevant for classification: We transform $X_O$ using $R(M_{\text{-}t}(X_O))$ and interpolate from
    $E(R(M_{\text{-}t}(X_O)))$ to $E(X_T)$ in the latent space given by $E$; as decoder we use $R'$.
    \item As the prior scenario but interpolating from $E(X_O)$ to $E(R(M_{\text{-}t}(X_T)))$.
\end{itemize}



\noindent\textbf{Setup.} We consider common DL methods. For the classifier we used different variants of VGG, i.e., VGG-11 and VGG-13, and ResNet-10. In particular, for the model $M$ to attack we focused on a $VGG-11$. We also used the adversarial samples on evaluation classifiers $M^{Ev} \in \{VGG-13, ResNet-10\}$ to see if adversarial samples are transferable. For VGG-11 we reconstructed after a ReLU unit associated with a conv layer. We used the second last layer corresponding to the last conv-layer of $VGG-11$ as target layer $L_{t}$, i.e., $t=10$. All encoders, i.e., $E$ and $M_{\text{-}t}$, share the same architecture, i.e., encoders are models $M_{\text{-}t}$ of a VGG-11 network and decoders $R$ and $R'$ are given by ClaDec\cite{sch21cla}. ClaDec use a standard decoder design relying on 5x5 deconvolutional layers. 

We employed two datasets namely Fashion-MNIST and MNIST. Fashion-MNIST consists of 70k 28x28 images of clothing stemming from 10 classes. MNIST of 60k digits objects in color; for both datasets, 10k samples are used for testing. As maximal difference of $||X_A - X_O||$ we used $\eta=15$ for Fashion-MNIST and $\eta=25$ for MNIST. The thresholds can be chosen arbitrarily: larger thresholds yield adversarial samples that more likely fool a classifier, but also look more different to the original sample $X_O$. As data-preprocessing, we scaled all images to 32x32, performed standardization and autoencoded all images using a separately trained autoencoder to ``smoothen'' outliers. For MNIST this makes very little difference, since autoencoders tend to almost perfectly reconstruct all samples. For Fashion-MNIST it has not impact for somewhat common samples, but it helps for outliers. For outliers, clothes with a seemingly random dotted pattern cannot be reconstructed well and are transformed to more common samples. For such outliers our method would otherwise fail to work, since the encoding and decoding using $AE=(E,R')$ even for the unmodified original sample $X_O$ would result in an error larger than the permitted threshold $\eta$, i.e., $||X_O-R'(E(X_O))||>\eta$.

We train all models for reconstruction using the Adam optimizer for 64 epochs, i.e., the autoencoder for data and the decoder $R$ from classifier representations $M_{\text{-}t}$. The classifiers to be attacked were trained using SGD for 64 epochs starting from a learning rate of 0.1 that was decayed twice by 0.1. We conducted 3 runs for each reported number, e.g., we trained all networks (classifiers, encoders, decoders) 5 times. We show both averages and standard deviations. The baseline performance on each dataset matches the state-of-the-art without data augmentation: for MNIST we achieved mean accuracy above 99\% and for Fashion-MNIST above 90\%. For each class $c$ we used all samples in the test data of classes not $c$ as original samples. For each original sample, the target sample $X_T$ was chosen randomly from samples of class $c$. Thus, we computed about 9k adversarial samples for each class (and each run). 



\begin{figure*}
\vspace{-6pt}
  \centering
  \includegraphics[width=\linewidth]{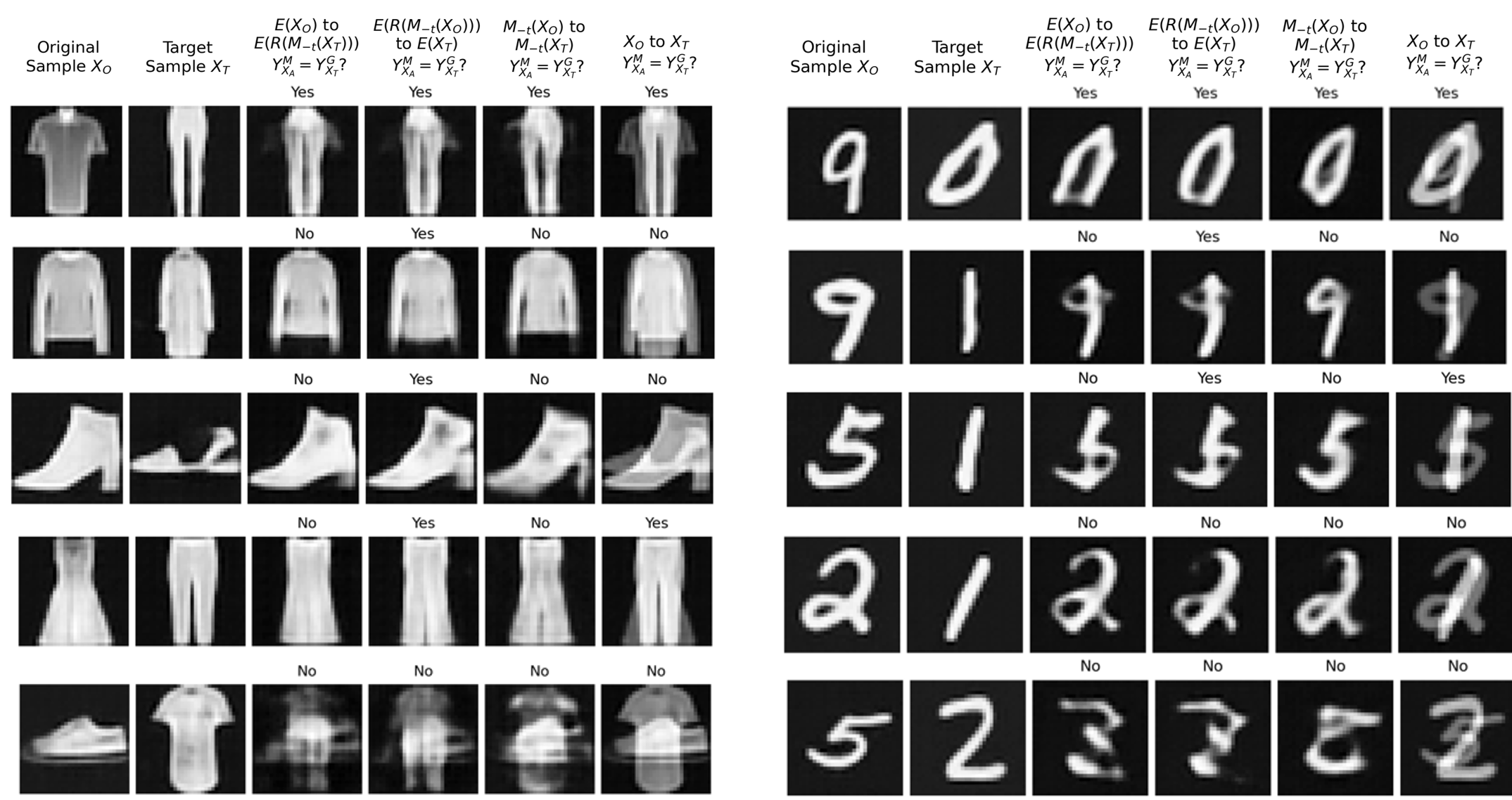}
  \caption{Original, target and adversarial samples for different en-/decodings and interpolation for Fashion-MNIST(left) and MNIST(right). Yes/No indicates, whether the model got fooled by $X_A$, i.e. it outputs the class of $X_T$ for $X_A$} \label{fig:safamn}
  \vspace{-12pt}
\end{figure*}


\subsection{Qualitative Results} \label{sec:qua} 
Fig.~\ref{fig:safamn} show adversarial samples for all four evaluated methods for interpolation. It can be observed that interpolation in image space ($X_O$ to $X_T$) is easily recognizable, i.e., both the original and target sample are well-recognizable when looking at the resulting adversarial sample. For example, in the last row (and last column) of each panel, it is easy to recognize the shoe and the T-shirt and digits 2 and 5. As such, the attack is easy to disguise. Methods employing encoding and decoding can yield non-interpretable images, as shown in the last row of each panel. This often happens if visual differences between the target and original samples are large. It also depends on the embedding, i.e., how classes are positioned in the encoded space. When comparing interpolation ($M_{\text{-}t}(X_O)$ to $M_{\text{-}t}(X_T)$) in latent space using the classifier $M_{\text{-}t}$ to those using the encoder $E$, it can be seen that the encodings using the classifier $M_{\text{-}t}$ tend to lead to non-desirable differences between the adversarial, target and original sample. For instance, in Fig.~\ref{fig:safamn} for the shoe in the 3rd row, it increases the height of the heel both the target sample and the original sample, which have no or lower heel size, i.e., the adversarial sample disagrees with both. Adversarial samples from ``$E(X_O)$ to $E(R(M_{\text{-}t}(X_T)))$'' and ``$E(R(M_{\text{-}t}(X_O)))$ to $E(X_T)$'' look similar, overall the former are somewhat better. This can be seen, for example, for the shoe in the 3rd column having a small black area (which also exists for the target sample, i.e., the sandal), and it has lower brightness in the upper part of the shoe compared to the original boot. This indicates a slight move towards the target sample, which suffices to classify the adversarial sample as the target sample, i.e., as a sandal. Thus, while differences between the reconstructions of the two methods appear more subtle, they are often sufficient to lead to different classification outcomes. However, it also becomes evident that some of the reconstructions appear to be instances of the other class. That is, they might be called ``forged samples'' that contain some but not very pronounced features of the original sample.  For example, the first row in Fig.~\ref{fig:safamn}. There are also samples containing elements of both target and original samples and seem somewhat ambiguous with respect to what class they belong to. For instance, the third row and 4th column for Fashion-MNIST depicts a sample that looks like a dress based on the upper part, but in the lower part it has elements of a pant, i.e., a dark area indicating a separation as for pant legs. For MNIST the second and third rows contain such samples, i.e., the nines (3rd and 4th column) do appear as nines but the loop of the "9" is not very pronounced and the samples can be easily mistaken as 1. Similarly, in the next column the 3rd and 4th column still look more like the original sample. Still, they contain elements of the target sample, i.e., a vertical line in the lower part, and  the top part of the 5 only has a very short horizontal bar. In conclusion, some generated samples differ only subtly from the original, some are mixtures containing elements of both, and some appear more like samples from the target class. The outcome depends on the maximal allowed perturbations and learnt embeddings, e.g., whether original and target classes are embedded near each other, which also depends on their visual similarity.


\subsection{Quantitative Results}

\begin{table*}[!htp]
 	\caption{Results for MNIST and FashionMNIST}  \label{tab:resFaMn} 
	\centering
 		\begin{tabular}{ |l|l| l|l|l|l|l| }\hline
 		 Dataset & Interpolation & \scriptsize{$||X_A-X_T||$} & \scriptsize{$||X_A-X_O||$} & \scriptsize{$Acc(M) \text{\tiny{(VGG-11)}})$} & \scriptsize{$Acc(M^{Ev})\text{\tiny{(VGG-13)}}$} & \scriptsize{$Acc(M^{Ev})\text{\tiny{(Res.-10)}}$}\\  \hline \hline

\multirow{4}{*}{MNIST}&$X_O$ to $X_T$ & 12.42\tiny{\text{$\pm$}1.25} & 24.73\tiny{\text{$\pm$}0.149} & 0.08\tiny{\text{$\pm$}0.073} & 0.11\tiny{\text{$\pm$}0.075}&0.09\tiny{\text{$\pm$}0.081}\\ \cline{2-7}
&$M_{-t}(X_O)$ to $M_{-t}(X_T)$ & 24.38\tiny{\text{$\pm$}1.71} & 24.71\tiny{\text{$\pm$}0.15} & 0.44\tiny{\text{$\pm$}0.117} & 0.41\tiny{\text{$\pm$}0.134}&0.42\tiny{\text{$\pm$}0.124}\\ \cline{2-7}
&$E(X_O)$ to $E(R(M_{-t}(X_T)))$ & 19.87\tiny{\text{$\pm$}1.794} & 24.85\tiny{\text{$\pm$}0.11} & 0.28\tiny{\text{$\pm$}0.081} & 0.26\tiny{\text{$\pm$}0.079}&0.27\tiny{\text{$\pm$}0.084}\\ \cline{2-7}
& $E(R(M_{-t}(X_O)))$ to $E(X_T)$ & 20.41\tiny{\text{$\pm$}1.837} & 24.73\tiny{\text{$\pm$}0.172} & 0.21\tiny{\text{$\pm$}0.078} & 0.2\tiny{\text{$\pm$}0.077}&0.2\tiny{\text{$\pm$}0.079}\\ \hline \hline

\multirow{4}{*}{\parbox{1.1cm}{Fashion-MNIST}}&$X_O$ to $X_T$ & 20.83\tiny{\text{$\pm$}1.317} & 14.95\tiny{\text{$\pm$}0.043} & 0.42\tiny{\text{$\pm$}0.14} & 0.44\tiny{\text{$\pm$}0.15}&0.41\tiny{\text{$\pm$}0.132}\\ \cline{2-7}
&$M_{-t}(X_O)$ to $M_{-t}(X_T)$ & 27.23\tiny{\text{$\pm$}1.44} & 14.84\tiny{\text{$\pm$}0.037} & 0.64\tiny{\text{$\pm$}0.052} & 0.62\tiny{\text{$\pm$}0.056}&0.62\tiny{\text{$\pm$}0.049}\\ \cline{2-7}
&$E(X_O)$ to $E(R(M_{-t}(X_T)))$ & 25.84\tiny{\text{$\pm$}1.436} & 14.85\tiny{\text{$\pm$}0.03} & 0.57\tiny{\text{$\pm$}0.059} & 0.58\tiny{\text{$\pm$}0.057}&0.56\tiny{\text{$\pm$}0.055}\\ \cline{2-7}
&$E(R(M_{-t}(X_O)))$ to $E(X_T)$ & 25.22\tiny{\text{$\pm$}1.365} & 14.92\tiny{\text{$\pm$}0.048} & 0.53\tiny{\text{$\pm$}0.065} & 0.53\tiny{\text{$\pm$}0.065}&0.51\tiny{\text{$\pm$}0.06}\\ \hline
 		\end{tabular}
 	
 	\vspace{-6pt}
 \end{table*}


 
From the quantitative results in Table~\ref{tab:resFaMn}, we can see that our samples are well-transferable between classifiers, i.e., an adversarial sample obtained using model $M$ also works on a different classifier without any changes to it. Furthermore, results are consistent for both datasets, i.e., the ordering of methods with respect to metrics, accuracy and difference to target sample $||X_A-X_T||$ is the same. We conducted statistical tests showing that differences between any two methods are significant at $p<0.001$ for accuracy and $||X_A-X_T||$ except the last two methods where we only have $p<0.01$.

We can also see that interpolating directly between the original and the target sample yields the lowest loss to the target sample, i.e., $||X_A-X_T||$, and the lowest correct classifications by $M$, i.e., $Acc(M)$. However, as shown in our qualitative evaluation, interpolations in the image space are often easily recognizable and, therefore, inadequate. Moreover, operating in the latent space given by layer $L_{t}$ (i.e., the encoder $M_{\text{-}t}$) yields poorest results among all methods (i.e., largest difference to the target sample $X_T$) and also leads to fewest samples $X_A$ being classified as $X_T$ (i.e., correct classification are highest). This is because reconstructions from  $M_{\text{-}t}$ are generally poor since the latent representations only accurately represent information relevant to classification. Thus, the constraint $||X_A-X_O||$ is violated without much movement towards the target sample $X_T$. For $X_T$ that can be considered rare or outliers it might happen that even the reconstruction $R(M_{\text{-}t}(X_T))$ of the latent representation $M_{\text{-}t}(X_T)$ of $X_T$ shows large differences to $X_T$. Using encodings $E(X_T)$ of the target sample leads to better results than using those of $R(M_{\text{-}t}(X_T))$ maintaining only classification relevant information. This is expected, since using $R(M_{\text{-}t}(X_T))$ we adjust $X_A$ towards a sample that differs from the proposed target $X_T$, i.e., we move towards the wrong ``target'', as also shown with examples in our qualitative evaluation.
Compared to other adversarial attacks, the model accuracy might seem fairly high, i.e., $Acc(M)$ is commonly above 30\% for adversarial samples. This is not unexpected, since often the original sample $X_O$ is difficult to transform to the target class, i.e., that of $X_T$. For instance, it is non-trivial to make a shoe appear like a T-shirt or turn a 2 into a 5 (see last row in Fig.~\ref{fig:safamn}). In such cases, the adversarial samples might be of low quality.

\section{Discussion and Future Work}
\label{sec:discussion}
We altered dense representations in an auto-encoded space and in a space resulting from layer activations of a classifier. Adversarial and original samples tend to exhibit visible differences that aim to deceive or nudge a human. The parameter $\eta$ allows controlling the degree of the differences. However, very small $\eta$ likely do not result in adversarial samples classified differently from the original sample for our technique based on linear interpolation. In future work, we might also consider attacks not based on a target sample, but directly manipulating layer encodings $X_O':=M_{\text{-}t}(X_O)$ using conventional attack methods (e.g.,~\cite{liu2019sensitivity}) to yield perturbations $p$. These attacks will result in misclassifications on the upper classifier $M_{t\text{-}}(X_O')$, i.e., $M_{t\text{-}}(X_O')\neq M_{t\text{-}}(X_O'+p)$, and adversarial samples $X_A:=R(X_O'+p)$ through reconstruction. Our optimization procedure using binary search converges to a local optimum, which could be improved by using multiple search intervals. 

To derive our results, we first preprocessed data using a separate autoencoder. Generally, our method (Fig.~\ref{fig:over}) using encoders and decoders introduces noise that reduces the possible addition of perturbation through optimization towards adversarial samples. Therefore, high-quality encoders and decoders and, in turn, sufficient training data for them are relevant and a limitation of our technique. While public datasets are available for many real-world objects, for more specialized datasets such as MRI images, this might not hold. 

One might also approach our problem as a two-step problem: Create a sample that fools a human and then alter this sample using conventional adversarial attacks. However, this defies one of our motivations, i.e., an adversarial sample should be ``realistic'' in the sense that it is a sample that might occur in normal usage. Additionally, joint optimization of both objectives possibly leads to better outcomes.

Our method requires complete knowledge of a DL model, which can be associated with `white-box' attackers. We observe that, in some domains, our assumptions are viable: for instance, many deep learning systems are trained on the ImageNet dataset~\cite{kornblith2019better}, which is publicly available~\cite{deng2009imagenet}. In these settings, even a `black-box' attacker can be successful because they can create a surrogate model and transfer the successful adversarial examples to the original model~\cite{demontis2019adversarial}. Our evaluation showed that our adversarial examples can be leveraged for similar strategies (§\ref{sec:evaluation}).





\section{Conclusions}
\label{sec:conclusions}
In this paper, we investigated adversarial attacks to construct adversarial samples that might present visible differences to a given (original) sample. Our adversarial samples should be similar to a target sample though they are still constrained to be similar to the original sample. Considering that existing systems rely on the cooperation of AI and humans, our adversarial samples will represent an attractive strategy for well-motivated attackers.



\bibliographystyle{IEEEtran.bst}
\bibliography{refs, references}

\begin{thebibliography}{10}
\providecommand{\url}[1]{#1}
\csname url@samestyle\endcsname
\providecommand{\newblock}{\relax}
\providecommand{\bibinfo}[2]{#2}
\providecommand{\BIBentrySTDinterwordspacing}{\spaceskip=0pt\relax}
\providecommand{\BIBentryALTinterwordstretchfactor}{4}
\providecommand{\BIBentryALTinterwordspacing}{\spaceskip=\fontdimen2\font plus
\BIBentryALTinterwordstretchfactor\fontdimen3\font minus
  \fontdimen4\font\relax}
\providecommand{\BIBforeignlanguage}[2]{{%
\expandafter\ifx\csname l@#1\endcsname\relax
\typeout{** WARNING: IEEEtran.bst: No hyphenation pattern has been}%
\typeout{** loaded for the language `#1'. Using the pattern for}%
\typeout{** the default language instead.}%
\else
\language=\csname l@#1\endcsname
\fi
#2}}
\providecommand{\BIBdecl}{\relax}
\BIBdecl

\bibitem{akh21}
N.~Akhtar, A.~Mian, N.~Kardan, and M.~Shah, ``Advances in adversarial attacks
  and defenses in computer vision: A survey,'' \emph{IEEE Access}, vol.~9, pp.
  155\,161--155\,196, 2021.

\bibitem{su2019one}
J.~Su, D.~V. Vargas, and K.~Sakurai, ``One pixel attack for fooling deep neural
  networks,'' \emph{IEEE Transactions on Evolutionary Computation}, vol.~23,
  no.~5, pp. 828--841, 2019.

\bibitem{var18}
A.~Varma, A.~Sarma, S.~Doshi, and R.~Nair, ``House price prediction using
  machine learning and neural networks,'' in \emph{2018 Second International
  Conference on Inventive Communication and Computational Technologies
  (ICICCT)}.\hskip 1em plus 0.5em minus 0.4em\relax IEEE, 2018, pp. 1936--1939.

\bibitem{sch20dec}
J.~Schneider, J.~Handali, M.~Vlachos, and C.~Meske, ``Deceptive {AI}
  explanations: Creation and detection,'' \emph{arXiv preprint
  arXiv:2001.07641}, 2020.

\bibitem{leo14}
N.~Leontiadis, T.~Moore, and N.~Christin, ``A nearly four-year longitudinal
  study of search-engine poisoning,'' in \emph{Proceedings of the 2014 ACM
  SIGSAC Conference on Computer and Communications Security}, 2014, pp.
  930--941.

\bibitem{jordan2015machine}
M.~I. Jordan and T.~M. Mitchell, ``{Machine learning: Trends, perspectives, and
  prospects},'' \emph{Science}, vol. 349, no. 6245, pp. 255--260, 2015.

\bibitem{lecun2015deep}
Y.~LeCun, Y.~Bengio, and G.~Hinton, ``Deep learning,'' \emph{Nature}, vol. 521,
  no. 7553, pp. 436--444, 2015.

\bibitem{biggio2013evasion}
B.~Biggio, I.~Corona, D.~Maiorca, B.~Nelson, N.~{\v{S}}rndi{\'c}, P.~Laskov,
  G.~Giacinto, and F.~Roli, ``Evasion attacks against machine learning at test
  time,'' in \emph{Joint European conference on machine learning and knowledge
  discovery in databases}.\hskip 1em plus 0.5em minus 0.4em\relax Springer,
  2013, pp. 387--402.

\bibitem{biggio2018wild}
B.~Biggio and F.~Roli, ``{Wild patterns: Ten years after the rise of
  adversarial machine learning},'' \emph{Elsevier Pattern Recognition},
  vol.~84, pp. 317--331, 2018.

\bibitem{papernot2018sok}
N.~Papernot, P.~McDaniel, A.~Sinha, and M.~P. Wellman, ``{SoK: Security and
  Privacy in Machine Learning},'' in \emph{Proc. IEEE European Symposium on
  Security and Privacy}, 2018, pp. 399--414.

\bibitem{pierazzi2020intriguing}
F.~Pierazzi, F.~Pendlebury, J.~Cortellazzi, and L.~Cavallaro, ``Intriguing
  properties of adversarial ml attacks in the problem space,'' in \emph{Proc.
  IEEE Symposium on Security and Privacy (SP)}, 2020, pp. 1332--1349.

\bibitem{cisse2017houdini}
M.~M. Cisse, Y.~Adi, N.~Neverova, and J.~Keshet, ``Houdini: Fooling deep
  structured visual and speech recognition models with adversarial examples,''
  \emph{Advances in neural information processing systems}, vol.~30, 2017.

\bibitem{apruzzese2019addressing}
G.~Apruzzese, M.~Colajanni, L.~Ferretti, and M.~Marchetti, ``Addressing
  adversarial attacks against security systems based on machine learning,'' in
  \emph{IEEE International Conference on Cyber Conflict}, vol. 900, 2019, pp.
  1--18.

\bibitem{liang2016cracking}
B.~Liang, M.~Su, W.~You, W.~Shi, and G.~Yang, ``Cracking classifiers for
  evasion: a case study on the google's phishing pages filter,'' in
  \emph{Proceedings of the 25th International Conference on World Wide Web},
  2016, pp. 345--356.

\bibitem{wu2020making}
Z.~Wu, S.-N. Lim, L.~S. Davis, and T.~Goldstein, ``Making an invisibility
  cloak: Real world adversarial attacks on object detectors,'' in
  \emph{European Conference on Computer Vision}, 2020, pp. 1--17.

\bibitem{madry2018towards}
A.~Madry, A.~Makelov, L.~Schmidt, D.~Tsipras, and A.~Vladu, ``Towards deep
  learning models resistant to adversarial attacks,'' in \emph{International
  Conference on Learning Representations}, 2018.

\bibitem{demontis2019adversarial}
A.~Demontis, M.~Melis, M.~Pintor, M.~Jagielski, B.~Biggio, A.~Oprea,
  C.~Nita-Rotaru, and F.~Roli, ``{Why do adversarial attacks transfer?
  Explaining transferability of evasion and poisoning attacks},'' in
  \emph{Proc. USENIX Security Symposium}, 2019, pp. 321--338.

\bibitem{papernot2017practical}
N.~Papernot, P.~McDaniel, I.~Goodfellow, S.~Jha, Z.~B. Celik, and A.~Swami,
  ``Practical black-box attacks against machine learning,'' in \emph{Proc. ACM
  Asia Computer and Communications Security Conference}, 2017, pp. 506--519.

\bibitem{carlini2018audio}
N.~Carlini and D.~Wagner, ``Audio adversarial examples: Targeted attacks on
  speech-to-text,'' in \emph{IEEE Security and Privacy Workshops (SPW)}, 2018,
  pp. 1--7.

\bibitem{zhang2021seat}
Z.~Zhang, Y.~Chen, and D.~Wagner, ``{SEAT: Similarity Encoder by Adversarial
  Training for Detecting Model Extraction Attack Queries},'' in
  \emph{Proceedings of the 14th ACM Workshop on Artificial Intelligence and
  Security}, 2021, pp. 37--48.

\bibitem{moosavi2016deepfool}
S.-M. Moosavi-Dezfooli, A.~Fawzi, and P.~Frossard, ``Deepfool: a simple and
  accurate method to fool deep neural networks,'' in \emph{Proceedings of the
  IEEE conference on computer vision and pattern recognition}, 2016, pp.
  2574--2582.

\bibitem{quiring2020adversarial}
E.~Quiring, D.~Klein, D.~Arp, M.~Johns, and K.~Rieck, ``Adversarial
  preprocessing: Understanding and preventing $\{$Image-Scaling$\}$ attacks in
  machine learning,'' in \emph{29th USENIX Security Symposium (USENIX Security
  20)}, 2020, pp. 1363--1380.

\bibitem{xiao2019seeing}
Q.~Xiao, Y.~Chen, C.~Shen, Y.~Chen, and K.~Li, ``Seeing is not believing:
  Camouflage attacks on image scaling algorithms,'' in \emph{28th USENIX
  Security Symposium (USENIX Security 19)}, 2019, pp. 443--460.

\bibitem{sch21cla}
J.~Schneider and M.~Vlachos, ``Explaining neural networks by decoding layer
  activations,'' in \emph{International Symposium on Intelligent Data
  Analysis}, 2021, pp. 63--75.

\bibitem{bhatt2020explainable}
U.~Bhatt, A.~Xiang, S.~Sharma, A.~Weller, A.~Taly, Y.~Jia, J.~Ghosh, R.~Puri,
  J.~M. Moura, and P.~Eckersley, ``Explainable machine learning in
  deployment,'' in \emph{Proc. ACM Conference on Fairness, Accountability, and
  Transparency}, 2020, pp. 648--657.

\bibitem{amich2021explanation}
A.~Amich and B.~Eshete, ``Explanation-guided diagnosis of machine learning
  evasion attacks,'' in \emph{International Conference on Security and Privacy
  in Communication Systems}, 2021, pp. 207--228.

\bibitem{sch22b}
J.~Schneider and M.~Vlachos, ``Explaining classifiers by constructing familiar
  concepts,'' \emph{arXiv preprint arXiv:2203.04109}, 2022.

\bibitem{hen21}
D.~Hendrycks, K.~Zhao, S.~Basart, J.~Steinhardt, and D.~Song, ``Natural
  adversarial examples,'' in \emph{Proceedings of the IEEE/CVF Conference on
  Computer Vision and Pattern Recognition}, 2021.

\bibitem{rob20}
A.~Robey, H.~Hassani, and G.~J. Pappas, ``Model-based robust deep learning:
  Generalizing to natural, out-of-distribution data,'' \emph{arXiv preprint
  arXiv:2005.10247}, 2020.

\bibitem{wan21}
Z.~Wang, Q.~She, and T.~E. Ward, ``Generative adversarial networks in computer
  vision: A survey and taxonomy,'' \emph{ACM Computing Surveys (CSUR)},
  vol.~54, no.~2, pp. 1--38, 2021.

\bibitem{schne22c}
M.~Basalla, J.~Schneider, and J.~Brocke, ``Creativity of deep learning:
  Conceptualization and assessment,'' in \emph{Proceedings of the 14th
  International Conference on Agents and Artificial Intelligence}, 2022.

\bibitem{fred15}
M.~Fredrikson, S.~Jha, and T.~Ristenpart, ``Model inversion attacks that
  exploit confidence information and basic countermeasures,'' in
  \emph{Proceedings of the 22nd ACM SIGSAC conference on computer and
  communications security}, 2015, pp. 1322--1333.

\bibitem{van20}
J.~van Doorenmalen and V.~Menkovski, ``Evaluation of cnn performance in
  semantically relevant latent spaces,'' in \emph{Int. Symposium on Intelligent
  Data Analysis}, 2020.

\bibitem{apruzzese2021modeling}
G.~Apruzzese, M.~Andreolini, L.~Ferretti, M.~Marchetti, and M.~Colajanni,
  ``Modeling realistic adversarial attacks against network intrusion detection
  systems,'' \emph{ACM Digital Threats: Research and Practice}, 2021.

\bibitem{shafahi2018poison}
A.~Shafahi, W.~R. Huang, M.~Najibi, O.~Suciu, C.~Studer, T.~Dumitras, and
  T.~Goldstein, ``Poison frogs! targeted clean-label poisoning attacks on
  neural networks,'' \emph{Advances in neural information processing systems},
  vol.~31, 2018.

\bibitem{liu2019sensitivity}
Y.~Liu, S.~Mao, X.~Mei, T.~Yang, and X.~Zhao, ``Sensitivity of adversarial
  perturbation in fast gradient sign method,'' in \emph{2019 IEEE Symposium
  Series on Computational Intelligence (SSCI)}.\hskip 1em plus 0.5em minus
  0.4em\relax IEEE, 2019, pp. 433--436.

\bibitem{kornblith2019better}
S.~Kornblith, J.~Shlens, and Q.~V. Le, ``{Do better Imagenet models transfer
  better?}'' in \emph{Proceedings of the IEEE/CVF Conference on Computer Vision
  and Pattern Recognition}, 2019, pp. 2661--2671.

\bibitem{deng2009imagenet}
J.~Deng, W.~Dong, R.~Socher, L.-J. Li, K.~Li, and L.~Fei-Fei, ``Imagenet: A
  large-scale hierarchical image database,'' in \emph{Proc. IEEE Conference on
  computer vision and pattern recognition}, 2009, pp. 248--255.

\end{thebibliography}

\end{document}